\documentclass{article}

\usepackage[dblblindworkshop, final]{neurips_2026}
\workshoptitle{Diffusion Language Models (DiffuLM)}

\usepackage[utf8]{inputenc}
\usepackage[T1]{fontenc}
\usepackage{hyperref}
\hypersetup{hidelinks}          
\usepackage{url}
\usepackage{booktabs}
\usepackage{amsfonts}
\usepackage{amsmath}
\usepackage{nicefrac}
\usepackage{microtype}
\usepackage{xcolor}
\usepackage{graphicx}
\usepackage{enumitem}

\title{The Hidden Cost of MoE Diffusion Decoding:\\ Profiling and Fixing the Sampler}

\author{%
  Harshit Samrat \\
  Indian Institute of Technology Madras \\
  \texttt{bs23b004@smail.iitm.ac.in} \\
  \And
  Ramakrishna Paramahamsa Kompella \\
  Indian Institute of Technology Madras \\
  \texttt{be21b029@smail.iitm.ac.in} \\
}

\begin{document}

\maketitle

\begin{abstract}
Masked diffusion language models (dLLMs) decode by iteratively denoising a full canvas of tokens, so every forward pass runs a per-token sampling and stopping pipeline over the entire \emph{canvas} $\times$ \emph{vocabulary} logits tensor. We profile single-user decoding of a 26B-parameter diffusion Mixture-of-Experts model on a bandwidth-bound edge accelerator (NVIDIA GB10) and report a result that is easy to miss: after weight quantisation, this logits pipeline---not the expert GEMMs---is the single largest consumer of decode time. Unlike autoregressive decoding, where sampling touches one position's logits per committed token, diffusion decoding processes $S{\times}$ more logit entries per output token, where $S$ is the number of denoising steps. We show this pipeline is largely redundant or fusable and remove it in two lossless tiers: a \emph{bit-exact} tier whose output is byte-identical to the reference decoder, and an \emph{accuracy-parity} tier using a fused single-pass entropy kernel and Gumbel-max sampling. Combined with a W4A16 expert kernel, the full stack reaches $2.4$--$2.7\times$ end-to-end speedup over \texttt{bfloat16} at unchanged benchmark accuracy on GSM8K, HumanEval, and MATH, on a single 273\,GB/s device. We confirm the finding on a second, independently built diffusion MoE (LLaDA2.0-mini: pipeline $28.8\%$ of decode, fused W4A16 $3.3\times$ at accuracy parity) and show it disappears on a \emph{dense} diffusion model ($3.6\%$)---the cost is specific to MoE diffusion, where quantisation makes the backbone cheap. Code and quantised weights are released.
\end{abstract}

\section{Introduction}

Masked diffusion language models \citep{austin2021d3pm,lou2024sedd,sahoo2024mdlm,nie2025llada} generate text by repeatedly denoising a fixed-length canvas: each forward pass predicts a distribution over the vocabulary for \emph{every} canvas position, commits a subset of positions according to a confidence criterion, and re-noises the rest for the next step. This differs fundamentally from autoregressive (AR) decoding, where each forward pass produces logits for a single next position. The difference is usually discussed in terms of parallelism and quality; we point out that it also reshapes \emph{where inference time goes}.

In AR decoding, the logits post-processing pipeline---temperature scaling, softmax, sampling, and any stopping heuristics---runs once per committed token, over a $[1,V]$ logits vector, and is negligible next to the backbone matrix multiplications. In diffusion decoding, the same pipeline runs on a $[C,V]$ tensor at \emph{every} denoising step, where $C$ is the canvas length and $V$ the vocabulary. For a model with $V{=}262{,}144$ and $C{=}256$, that is a $67$M-element, $268$\,MiB tensor touched several times per step, and there are many steps per generated block. Per output token, the logits pipeline processes roughly $S\times$ more entries than AR, where $S$ is the number of denoising steps (Section~\ref{sec:profile}). This is a structural cost of diffusion decoding, and it has gone largely unmeasured because dLLMs are rarely served under latency pressure.

We study this cost concretely on \textbf{DiffusionGemma-26B-A4B}, an open-weights masked-diffusion Mixture-of-Experts (MoE) model (128 experts, top-8 routing, 30 layers, hidden size 2816), decoded single-user on an \textbf{NVIDIA GB10} (DGX Spark: 273\,GB/s unified LPDDR, 119\,GB, one user, batch size 1)---a regime where decoding is bandwidth-bound and every wasted pass is visible in wall-clock latency. Our contributions:

\begin{itemize}[leftmargin=1.2em,itemsep=1pt,topsep=1pt]
\item \textbf{A profiling result.} After standard W4A16 expert quantisation removes the dominant weight-bandwidth cost, the logits/sampling pipeline becomes the \emph{largest single bucket} of decode time ($37\%$ averaged over 40 queries), exceeding the expert GEMMs ($26\%$) on every query we measured. We give the $O(S)$-per-token analysis that explains why (Section~\ref{sec:profile}).
\item \textbf{Two lossless optimisations} that target this pipeline. These apply known techniques (deduplication, online softmax, Gumbel-max); the profiling observation is the paper's main contribution, and the optimisations are what it implies (Section~\ref{sec:opt}): (i) a \emph{bit-exact} elimination of a redundant entropy recomputation, with byte-identical output; (ii) an \emph{accuracy-parity} fused single-pass entropy kernel and Gumbel-max sampler that replace a multi-pass softmax/multinomial chain.
\item \textbf{An integrated edge-serving stack} for a 26B diffusion MoE on one 273\,GB/s device, reaching $2.4$--$2.7\times$ over \texttt{bfloat16} at unchanged accuracy, with a strictly bit-exact sub-tier (Section~\ref{sec:results}). Weights and code are released.\footnote{Quantised model: \url{https://huggingface.co/GoedelMachines/diffusiongemma-26B-A4B-w4a16}}
\item \textbf{Generality.} We reproduce the profiling result and the fused-W4A16 speedup ($3.3\times$, accuracy parity) on a second diffusion MoE, and show a dense diffusion model is the counterexample---so the pipeline cost is a property of MoE diffusion, and its \emph{fusable} portion depends on the sampling algorithm (Section~\ref{sec:general}).
\end{itemize}

\section{Where Does Diffusion Decode Time Go?}
\label{sec:profile}

\paragraph{Setup.} We serve DiffusionGemma-26B-A4B with its reference decoding loop (\texttt{model.generate}) and quantise the MoE experts to asymmetric 4-bit weights with 16-bit activations (W4A16, group sizes 128/64), reducing expert storage from 85.1 to 21.3\,GiB. Quantisation is applied through a single fused grouped-GEMM Triton kernel~\citep{tillet2019triton} that serves all 128 experts in one launch with a deterministic (atomic-free) combine; this is standard practice~\citep{frantar2023gptq,lin2024awq} and is not our contribution. It is, rather, the tool that exposes the finding: once expert weight traffic is cut by $4\times$, whatever remains as the bottleneck is what matters for diffusion serving.

\paragraph{The pipeline dominates.} Table~\ref{tab:profile} breaks down quantised decode time over 20 GSM8K and 20 HumanEval queries (canvas $C{=}256$), measured with CUDA-event hooks around each submodule. The experts, despite being the bulk of the parameters, take only about a quarter of the time. The largest single bucket is the logits/sampling pipeline together with its host-side glue: $34$--$40\%$ on average, more than the experts on all 40 queries (minimum $31.2\%$ against an expert maximum of $26.6\%$), and roughly double attention. One GSM8K run hit a host-side stall ($456$\,ms per forward against a median of $219$); without it, the GSM8K pipeline share is $38.2\pm2.4\%$. The share is lower on HumanEval because its queries take half as many forwards ($5.8$ versus $11.6$), so the one-off prompt encoding claims more of the total. An operator-level trace over a representative GSM8K run attributes this bucket to a small number of full-vocabulary passes over the $[256,262144]$ logits tensor: the token-entropy computation used by the acceptance and stopping criteria ($\approx 29$\,ms/forward), the softmax feeding multinomial sampling ($\approx 20$\,ms), the multinomial draw ($\approx 10$\,ms), and an argmax ($\approx 2.5$\,ms). Critically, the entropy is computed \emph{twice} per step---once inside the token-acceptance sampler and once inside the stopping criterion---on the same tensor.

\begin{table}[t]
\centering
\small
\caption{Decode-time breakdown of the W4A16 model (GB10, canvas 256): share of wall-clock time per bucket, mean $\pm$ std over $n{=}20$ queries per benchmark, and mean ms per forward over all 40. Buckets are disjoint. The logits/sampling pipeline (with host glue) is the largest bucket on every query.}
\label{tab:profile}
\setlength{\tabcolsep}{5pt}
\begin{tabular}{lccr}
\toprule
Component & GSM8K (\%) & HumanEval (\%) & ms/fwd \\
\midrule
\textbf{Logits / sampling pipeline + host glue} & \textbf{39.8\,$\pm$\,7.4} & \textbf{34.4\,$\pm$\,2.7} & \textbf{87.0} \\
Expert GEMMs (W4A16 MoE) & 25.3\,$\pm$\,3.0 & 25.8\,$\pm$\,0.5 & 57.9 \\
Attention & 16.3\,$\pm$\,1.9 & 17.1\,$\pm$\,0.4 & 37.9 \\
Encoder (prompt KV, amortised) & 5.9\,$\pm$\,2.1 & 9.8\,$\pm$\,2.3 & 17.9 \\
RMSNorms & 5.9\,$\pm$\,0.7 & 5.9\,$\pm$\,0.2 & 13.3 \\
Shared MLP & 3.5\,$\pm$\,0.4 & 3.8\,$\pm$\,0.1 & 8.3 \\
LM head & 3.3\,$\pm$\,0.4 & 3.2\,$\pm$\,0.1 & 7.4 \\
\bottomrule
\end{tabular}
\end{table}

\paragraph{Why: an $O(S)$ per-token asymmetry.} Consider generating a block of $C$ tokens. An AR decoder runs $C$ forward passes, each sampling over a $[1,V]$ vector, for $C\cdot V$ logit entries of pipeline work, i.e.\ $V$ per output token. A diffusion decoder runs $S$ denoising steps, each running the full pipeline over the $[C,V]$ canvas, for $S\cdot C\cdot V$ entries, i.e.\ $S\cdot V$ per output token. The logits pipeline is therefore $\approx S\times$ more expensive per token under diffusion, where $S$ is the step count. In our runs $S$ is roughly $5$--$12$ (we observe $11.8$ and $5.7$ forward passes per query on GSM8K and HumanEval respectively), so the pipeline that is free in AR becomes a first-order serving cost. This asymmetry is intrinsic to canvas-based diffusion decoding and independent of the specific acceptance rule. The per-token argument explains why the pipeline costs more than in AR decoding; its \emph{share} of each forward comes from a second asymmetry. In a bandwidth-bound forward the expert weights are read once and shared by all $C$ canvas positions, so backbone time is set mostly by weight bytes, while the logits pipeline touches $C\cdot V$ elements that nothing amortises. Quantisation cuts only the first term, which is why the pipeline overtakes the experts after W4A16.

\section{Lossless Pipeline Optimisations}
\label{sec:opt}

We define two levels of ``lossless.'' \emph{Bit-exact}: output token IDs and per-query step counts are identical to the reference decoder for every query. \emph{Accuracy-parity}: benchmark accuracy is statistically unchanged, though individual token trajectories may differ (e.g.\ a different but equivalent random stream). We target the pipeline of Section~\ref{sec:profile} at both levels.

\paragraph{Tier 1 (bit-exact): remove the duplicate entropy.} The reference loop computes the per-position token entropy twice on the same logits---once to decide which tokens to accept, once to decide whether to stop. We compute it once and pass it through. Because the second consumer now receives the exact tensor the first produced, and floating-point reductions are run-to-run deterministic for fixed shape, dtype, and device, the outputs are unchanged bit-for-bit. The only subtlety is that when finished sequences rebuild the logits tensor, the cached entropy no longer describes it; that branch falls back to recomputation. We verified byte-identical text and step counts on all queries (Section~\ref{sec:results}), and have upstreamed this optimisation to the model's reference framework.

\paragraph{Tier 2 (accuracy-parity): fuse the entropy and sampling passes.} The remaining pipeline is several separate full-vocabulary kernels. We replace them with two fused Triton kernels.

\emph{(a) One-pass entropy.} The entropy $H=\log\!\sum_j e^{x_j} - \frac{\sum_j x_j e^{x_j}}{\sum_j e^{x_j}}$ is computed in a single streaming pass using an online-softmax accumulation of the running max, $\sum e$, and $\sum x e$, replacing a five-kernel chain (a $\approx 16\times$ reduction in this op's cost) at $<10^{-6}$ relative error.

\emph{(b) Gumbel-max sampling.} Categorical sampling as \texttt{softmax}~$\rightarrow$~\texttt{multinomial} materialises a $268$\,MiB probability tensor and makes several passes. The Gumbel-max identity~\citep{gumbel1954,maddison2014astar,jang2017gumbel}, $\arg\max_j (x_j + g_j)$ with $g_j\sim\mathrm{Gumbel}(0,1)$, samples \emph{exactly} the same categorical distribution as the softmax in one reduction. We generate the Gumbel noise inline with a counter-based RNG so it is never stored, and recover the greedy argmax (needed by the stopping criterion) from the same pass at no extra cost. Sampling drops from $\approx 32$\,ms to $\approx 1.2$\,ms per forward. Because the RNG stream differs from the reference \texttt{multinomial}, this tier is accuracy-parity rather than bit-exact: the distribution is identical, the specific draws are not.

A third, orthogonal fused-RMSNorm kernel (folding the reference $\approx 6$-kernel normalisation into one launch per norm) contributes to the full ``turbo'' configuration. All three are gated on the definitions above.

\section{Results}
\label{sec:results}

\paragraph{End-to-end.} Table~\ref{tab:main} reports the full stack against a \texttt{bfloat16} baseline, both arms run in the same process on the reference \texttt{model.generate} loop, GSM8K/HumanEval/MATH at $n{=}100$ each. The bit-exact tier alone reaches $1.8$--$2.0\times$ while producing byte-identical output to the quantised reference decoder; the accuracy-parity ``turbo'' configuration reaches $2.4$--$2.7\times$, with an accuracy delta of $+1/{-}1/{-}1$ across the 300 turbo questions. At $n{=}100$ the 95\% binomial confidence interval on each pass rate has a half-width of $4$--$7$ points ($\pm3.8$ at $96\%$, $\pm6.6$ at $87\%$), so these differences are well inside sampling error; we do not claim equality beyond that resolution. Throughput is measured as decoded output tokens per wall-clock second (diffusion commits multiple tokens per forward).

\begin{table}[t]
\centering
\small
\caption{End-to-end decode throughput on GB10, single user, reference \texttt{model.generate}; each cell is \emph{tok/s} (\emph{speedup} over the \texttt{bfloat16} baseline). Quantising the experts alone gives $\sim\!1.6$--$1.7\times$; the two logits-pipeline tiers of Section~\ref{sec:opt} contribute the rest, reaching $2.4$--$2.7\times$. \textbf{Accuracy is preserved at every step}: on a same-session $n{=}100$ comparison, \texttt{bfloat16} scores $96/89/88$ and turbo scores $97/88/87$ (GSM8K\,/\,HumanEval\,/\,MATH, exact-match or pass@1); the bit-exact tier is token-identical to the W4A16 decoder, and W4A16 quantisation is accuracy-neutral versus \texttt{bfloat16} ($96{\to}96$, $87{\to}87$ on GSM8K\,/\,MATH).}
\label{tab:main}
\setlength{\tabcolsep}{6pt}
\begin{tabular}{lccc}
\toprule
Configuration & GSM8K & HumanEval & MATH \\
\midrule
\texttt{bfloat16} baseline & 70.4 & 84.3 & 69.7 \\
W4A16 experts (reference loop) & 118.1\,(1.68$\times$) & 140.6\,(1.67$\times$) & 110.2\,(1.58$\times$) \\
\quad + bit-exact pipeline$^\dagger$ & 134.4\,(1.91$\times$) & 167.6\,(1.99$\times$) & 123.9\,(1.78$\times$) \\
\quad + turbo pipeline & \textbf{186.2\,(2.64$\times$)} & \textbf{224.9\,(2.67$\times$)} & \textbf{168.4\,(2.42$\times$)} \\
\bottomrule
\end{tabular}
\\[3pt]
\parbox{\linewidth}{\footnotesize $^\dagger$Bit-exact outputs are byte-identical to the W4A16 experts-only decoder (token- and step-count-identical, verified $20/20$; Table~\ref{tab:ladder}). The \texttt{bfloat16} and turbo rows are one same-process $n{=}100$ run; the W4A16-experts and bit-exact rows are separate runs whose \texttt{bfloat16} baseline reproduced to within $1.5\%$, so all speedups are throughput ratios against the common baseline shown.}
\end{table}

\paragraph{Attribution.} Table~\ref{tab:ladder} isolates each pipeline optimisation on GSM8K ($n{=}20$, same model load). Starting from the quantised model at $117.5$ tok/s, the bit-exact entropy dedup adds $+13\%$, the fused entropy and Gumbel-max sampler take it to $165.1$, and fused norms to $200.3$---a cumulative $1.7\times$ on top of quantisation, entirely from the logits/normalisation pipeline rather than the backbone. The same stack contributes $1.58\times$ in Table~\ref{tab:main} ($118.1\to186.2$). The two numbers come from different query sets (the first 20 versus the first 100 GSM8K problems) and separate runs; per-query throughput varies by more than $2\times$ with generation length and step count, so we read the pipeline tiers' contribution as roughly $1.6$--$1.7\times$ rather than a single value.

\begin{table}[t]
\centering
\small
\caption{Attribution of pipeline optimisations, GSM8K $n{=}20$, same load. Accuracy is stable ($19$--$20/20$) throughout. The bit-exact tier's outputs are byte-identical to the quantised reference (texts and step counts match $20/20$ on GSM8K and HumanEval).}
\label{tab:ladder}
\begin{tabular}{lcc}
\toprule
Configuration & tok/s & tier \\
\midrule
W4A16 experts (reference loop) & 117.5 & --- \\
\;+ entropy dedup & 132.4 & bit-exact \\
\;+ fused entropy \& Gumbel-max & 165.1 & accuracy-parity \\
\;+ fused RMSNorm (turbo) & 200.3 & accuracy-parity \\
\bottomrule
\end{tabular}
\end{table}

\paragraph{Bit-exactness.} On both GSM8K and HumanEval ($n{=}20$), the bit-exact tier produces token-identical output and identical per-query step counts to the quantised reference decoder ($20/20$), confirming it changes speed without changing the model's behaviour.

\section{Generality Across MoE Diffusion Models}
\label{sec:general}

To test whether these findings are specific to DiffusionGemma, we profile two further masked-diffusion LMs on the same device: \textbf{LLaDA2.0-mini}~\citep{llada2}, an independently built diffusion MoE (256 experts, top-8, 20 layers, $V{=}157{,}184$), and \textbf{LLaDA-8B}~\citep{nie2025llada}, a \emph{dense} diffusion model ($V{=}126{,}464$). Table~\ref{tab:general} shows the logits-pipeline share of decode: it is a first-order cost on both MoE models ($28.8$--$37.1\%$) but negligible on the dense model ($3.6\%$). The mechanism is the same as Section~\ref{sec:profile}: on a dense model the per-canvas projections are compute-bound GEMMs and the backbone dominates; MoE routing makes each expert bandwidth-bound, so once experts are cheap the pipeline surfaces. Vocabulary size is a confound we cannot fully remove with three models: LLaDA-8B's $V$ is about half of DiffusionGemma's. A smaller $V$ alone would shrink the pipeline's share by at most about half, not ten-fold, and LLaDA2.0-mini ($V{=}157$K, close to LLaDA-8B's) still shows a $28.8\%$ share, though under a heavier top-$p$ sampler than LLaDA-8B's greedy loop; we therefore attribute most of the gap to the backbone regime, with vocabulary and sampler as secondary factors. A controlled sweep over $V$ at fixed architecture is left to future work.

\begin{table}[t]
\centering
\small
\setlength{\tabcolsep}{6pt}
\caption{Logits-pipeline share of per-forward decode across three masked-diffusion LMs (GB10). It is first-order on MoE models, negligible on the dense one. \emph{Below:} porting the fused W4A16 MoE kernel to LLaDA2.0-mini gives $3.3$--$3.4\times$ over its \texttt{bfloat16} reference at accuracy parity ($n{=}50$; $\Delta$ within run-to-run sampling noise).}
\label{tab:general}
\begin{tabular}{lcc}
\toprule
Model & Arch. & Pipeline share \\
\midrule
DiffusionGemma-26B (W4) & MoE & 37.1\% \\
LLaDA2.0-mini (bf16) & MoE & 28.8\% \\
LLaDA-8B (bf16) & dense & 3.6\% \\
\bottomrule
\end{tabular}
\hfill
\begin{tabular}{lcc}
\toprule
LLaDA2.0-mini & bf16 & W4A16 \\
\midrule
GSM8K & 38/50 @ 18.7 & 41/50 @ 61.8 (3.3$\times$) \\
HumanEval & 38/50 @ 57.3 & 35/50 @ 191.9 (3.4$\times$) \\
MATH-500 & 28/50 @ 20.3 & 27/50 @ 68.7 (3.4$\times$) \\
\bottomrule
\end{tabular}
\end{table}

The fused W4A16 MoE kernel (grouped GEMM, SiLU epilogue) transfers directly: on LLaDA2.0-mini it is $3.3$--$3.4\times$ faster than the reference per-expert loop at accuracy parity across GSM8K, HumanEval, and MATH-500 (Table~\ref{tab:general}, lower), reproducing the DiffusionGemma result on a second, independently trained MoE. The \emph{pipeline} optimisations, however, transfer only in proportion to the pipeline's composition. DiffusionGemma's loop is sort-free (duplicated entropy plus multinomial), so it fuses cleanly for the $+40\%$ of Section~\ref{sec:opt}. LLaDA2.0-mini instead top-$p$-filters over a $157$K vocabulary every step; Gumbel-max removes its multinomial but the top-$p$ sort remains, so the accuracy-neutral variant nets $\approx\!0$. In a same-session run on MATH-500 ($n{=}50$), W4A16 alone gives $26/50$ at $67.8$ tok/s and the Gumbel-max sampler with top-$p$ kept gives $27/50$ at $71.3$; across two runs the difference between them ranged from $-5\%$ to $+5\%$, which we read as no reliable gain. Dropping top-$p$ removes the sort but bought no reliable speed ($70.5$ tok/s) and scored lower in both runs ($24/50$ and $23/50$, versus $26$--$27/50$ for every other arm), so we do not recommend it. The pipeline is first-order across MoE diffusion, but which of it is fusable depends on the sampling algorithm.

\section{Related Work}

Work on accelerating diffusion LLMs has mostly reduced the number or cost of forward passes: approximate KV caching and confidence-aware parallel decoding~\citep{wu2025fastdllm,ma2025dkvcache}. These leave the per-step logits pipeline untouched and are complementary to ours. Closest in diagnosis, \citet{lou2026npu} also find that sampling takes a large share of dLLM inference latency (up to $71\%$ on an A6000) and propose a hardware remedy; we reach a compatible conclusion on commodity hardware and address it in software. Fused sampling kernels are standard in autoregressive serving frameworks such as FlashInfer and vLLM~\citep{ye2025flashinfer,kwon2023vllm}, and FlashSampling~\citep{ruiz2026flashsampling} likewise uses Gumbel-max for exact single-pass sampling. Those systems target autoregressive decoding, where sampling covers one position per step; our point is that diffusion decoding multiplies this cost by the canvas and step count, which makes the same techniques first-order there.

\section{Discussion and Limitations}

The takeaway is not any single kernel but where to look: in masked-diffusion serving, the logits pipeline is a first-order cost that grows with the number of denoising steps, and much of it is redundant or fusable without touching model behaviour. Optimisations that are pointless for AR serving---deduplicating a full-vocabulary reduction, one-pass entropy, single-reduction sampling---pay back in proportion to $S$ here.

Our study spans three diffusion LMs but one device and single-user, batch-size-1 decoding; the $O(S)$ analysis is general, but the absolute breakdown will shift on datacenter GPUs with more compute headroom and on models with smaller vocabularies. Batching is the most consequential open question: it amortises expert weight reads across requests, but the logits pipeline grows linearly with batch size, so we expect its share to stay first-order or grow under batching, though we have not measured it. We measured the eager execution path. We attempted a \texttt{torch.compile} baseline, but the reference loop's compiled path recompiles for every prompt length and exceeded the compiler's recompilation limit on our benchmark mix, so we report no compiled numbers; a compiled path would fuse more of the redundant work and shrink the bit-exact tier's margin, while the accuracy-parity kernels remove passes a compiler cannot. Finally, the accuracy-parity tier is lossless in distribution and in benchmark accuracy but not bit-exact by construction; we keep the two tiers separate so that applications requiring reproducible sampling can use the bit-exact path. We hope the profiling result encourages dLLM serving work to treat the sampling pipeline as a primary target rather than an afterthought.

\bibliographystyle{plainnat}
\bibliography{refs}

\end{document}